\documentclass[10pt,twocolumn,letterpaper]{article}
\usepackage[left=2.5cm, right=2.5cm, top=2.5cm,bottom=2.5cm]{geometry}

\usepackage{times}
\usepackage{epsfig}
\usepackage{graphicx}
\usepackage{amsmath}
\usepackage{multirow}
\usepackage{amssymb}
\newcommand{\dummyfig}[1]{
  \centering
  \fbox{
    \begin{minipage}[c][0.2\textheight][c]{0.45\textwidth}
      \centering{#1}
    \end{minipage}
  }
}
\newcommand{\largedummyfig}[1]{
  \centering
  \fbox{
    \begin{minipage}[c][0.2\textheight][c]{\textwidth}
      \centering{#1}
    \end{minipage}
  }
}
\usepackage{color}
\usepackage[table,xcdraw]{xcolor}
 \usepackage{booktabs}
\usepackage{algorithm}
\usepackage[noend]{algpseudocode}
\usepackage{bbm}

\usepackage[pagebackref=true,breaklinks=true,colorlinks,bookmarks=false]{hyperref}

\usepackage{authblk}

\begin{document}

\title{What can I do here? Leveraging Deep 3D saliency and geometry \\ for fast and scalable multiple affordance detection}
\author{Eduardo Ruiz and Walterio Mayol-Cuevas \\
Department of Computer Science\\
University of Bristol, UK\\
{\tt\small \{er13827,wmayol\}@bristol.ac.uk}}

\date{}
\maketitle


\begin{abstract}
This paper develops and evaluates a novel method that allows for the detection of affordances in a scalable and multiple-instance manner on visually recovered pointclouds. Our approach\footnote{Code available at :\url{https://github.com/eduard626/deep-interaction-tensor}} has many advantages over alternative  methods, as it is based on highly parallelizable, one-shot learning that is fast in commodity hardware. The approach is hybrid in that it uses a geometric representation together with a state-of-the-art deep learning method capable of identifying 3D scene saliency. The geometric component allows for a compact and efficient representation, boosting the performance of the deep network architecture which proved insufficient on its own.
Moreover, our approach allows not only to predict whether an input scene affords or not the interactions, but also the pose of the objects that allow these interactions to take place. 
Our predictions align well with crowd-sourced human judgment as they are preferred with 87\% probability, show high rates of improvement with almost four times (4x) better performance over a deep learning-only baseline and are seven times (7x) faster than previous art.
\end{abstract}

\section{Introduction}

Vision emerged to function and operate in the 3D world, and perhaps its most fundamental question is {\it what is it actually for?} 
The early propositions to computationally address what Vision is for by Marr \cite{Marr1982}, have set a path to aim, and often end with, the recovery of a representation about the environment.
\begin{figure}[ht]
        \centering
        \IfFileExists{./first_image_scannet2.pdf}{\includegraphics[width=0.48\textwidth]{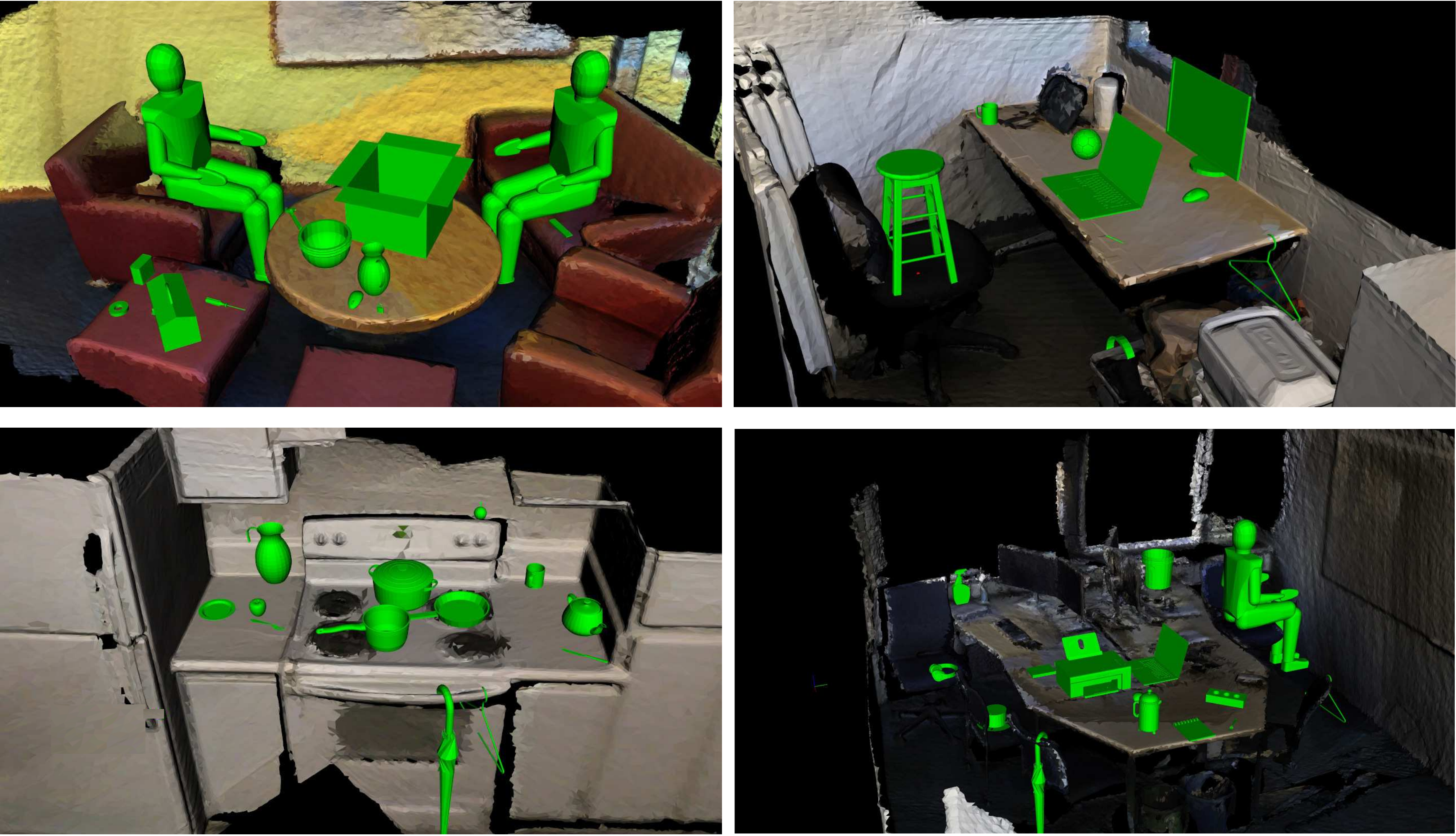}}{\dummyfig{Prediction examples on ScanNet}}
      \caption{Plausible scenes generated with interaction affordances (green objects and their poses) determined by our hybrid deep-geometric method. We determine over 80 affordances simultaneously in real-time on never before seen RGBD scenes.
      }
        \label{fig: first_image_examples}
\vspace{-2mm}
\end{figure}

\begin{figure*}[ht]
        \centering
        \IfFileExists{./first_big_figure4.pdf}{\includegraphics[width=0.98\textwidth]{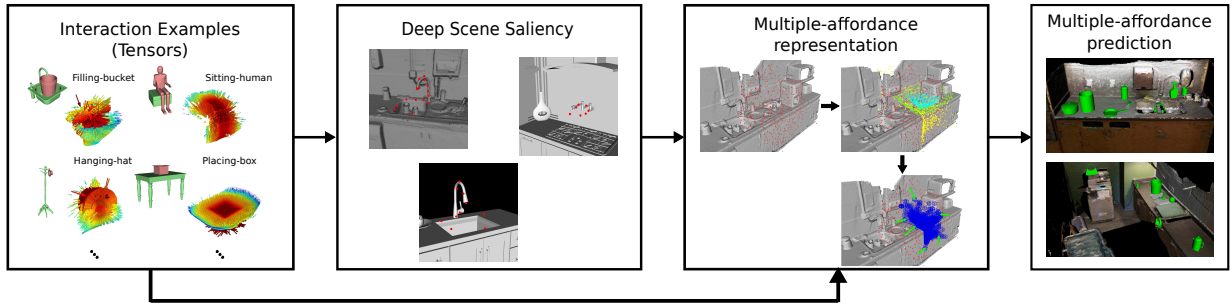}}{\largedummyfig{Overview of method}}
      \caption{We leverage the ability of the interaction tensor to generate single-affordance predictions from a single example. Then, using a deep neural network we learn 3D salient locations in order to construct and optimize a multiple-affordance descriptor. This allows us to answer questions such as “What can be afforded here?" on never-seen before scenes. Please see video material for further information.}
        \label{fig: method}
\vspace{-2mm}
\end{figure*}

However earlier, the view posed by J.J. Gibson \cite{gibson1977}, called for a visual perception that is there to rather help the perceiving agent to interact with the world. Through the coining of the term {\it affordance}, visual perception is described as a process to understand what can be done where. Such representation of the world is immediately useful, as by definition, it is one that already takes into account what the agent is capable of ---The affordance is about the interaction between object and scene and not about either of these in isolation.

In contrast, current efforts in visual affordance determination, starkly favor strategies that distance themselves from the agent's capabilities. Many of the approaches for affordance learning rely on object detection, segmentation or classification to then map functionality to specific object classes and from this to affordances \cite{AffordanceNet18,Chuang2018}. We believe that the learning of an affordance should not be about the classification of all possible object-instances (and scenes), or assigning arbitrary labels to them ---since a cup is not only for drinking but also a paperweight, or even a tool to build sand castles. Affordance determination is less about learning the label and more about capturing the intended {\it interaction}. Understanding the world via affordances is thus, arguably, a primary task for visual perception in the 3D world.

In this work, we develop a novel scalable, multiple-affordance approach that is capable of operating at high frame-rates in commodity desktop hardware (Figs \ref{fig: first_image_examples} and \ref{fig: method}). This approach leverages state-of-the-art deep learning methods for 3D data such as \cite{qi2017plus}, but also takes advantage of the one-shot learning approach for geometric affordances presented in \cite{Ruiz2018}. Similarly to the latter, our approach is agnostic to the "identity" of objects in the input scene. Furthermore, the method allows not only to predict whether a point in the scene affords or not the intended interactions but also how it is affordable. In other words, our approach provides the pose of the objects that allow the predicted interactions to take place. Fig. \ref{fig: method} depicts a general overview of our method.

We demonstrate results of our predictions on multiple RGBD scenes from public datasets, and include validation from human judgments via mechanical turk. Our approach is straight-forward yet results indicate we can detect the many dozens of affordances we evaluate (80+) in multiple previously-unseen settings. Results indicate high precision and input from crowd-sourced human judgment also shows how aligned our method is with what people expects to be able to afford at a given place.

The paper is organized as follows. Section \ref{sec:relatedwork} describes most common approaches studied to date in affordance detection as well as the state-of-the-art methods for learning on 3D data. Sections \ref{sec:affordancedetection} and \ref{section:saliency} describe in detail our approach before we discuss evaluation and results in Section \ref{section:evaluation}. Finally, Section \ref{conclusions} presents our conclusions and final remarks.

\section{Related work}
\label{sec:relatedwork}
\textbf{Affordance learning} Affordance detection has been studied in recent years in computer vision and robotics.

In the Robotics community the favored approach has been the representation and learning of actions. Mainly to predict consequences of actions over a set of objects \cite{Min2016,Mar2015}; or to learn to assist humans in every day task \cite{saponaro2013,pandey2013,pieropan2013,koppula2014,srikantha2014,chan2014,jiang2014}. These approaches use visual features describing shape, color, size and relative distances to capture object properties and effects. 

In Computer Vision, work has been done using static imagery, where the affordance or interaction are provided as a label rather than demonstrated. The works proposed in \cite{desai2013,srikantha2014,Than2017,ye2017,AffordanceNet18} are 
based on labeled 2D images to predict functional regions or attributes on every day objects. Approaches performing semantic reasoning from 2D images such as \cite{Zhu2014,Chao2015,Chuang2018} include human context to build knowledge representations useful 
for deciding on possible actions. Yet another body of research is the one exploiting 3D information to learn and predict affordances of objects in the environment. Affordances such as rollable, containment or sittable are studied in \cite{aldoma2012, hinkle2013} using simulations on 3D CAD models. In \cite{kim2014,Myers15,Nguyen2016} geometric features on RGB-D images are used to predict affordances such as pushable, liftable, graspable, support, cut or contain in a pixel-wise manner. Works such as \cite{Gupta2011,jiang2014,Piyathilak2015,Roy2016,wang2017} predict human poses or locations suitable for human activities such as sitting, walking or laying-down in indoor scenes.

\textbf{Learning on 3D data}  The interest in learning from 3D data in Computer Vision is clear due to leveraging labelling for free, increased training numbers or the reality of having to operate in the 3D world. But only recently, Convolutional Neural Networks (CNN) have started to work with this type of input aiming to catch up with the progress achieved on 2D imagery. Examples of such works are \cite{voxnet2015,Zhirong15CVPR,wang2015voting,qi2016volumetric,li2016fpnn,Brock2016,KlokovL17,Riegler2017OctNet}, which use voxelization to transform irregular data into occupancy grids that allow 3D convolutions to be applied. Even more recently, deep learning algorithms that work on point representations (pointclouds) have been proposed. Among these latest approaches are \cite{qi2016pointnet,yi2017syncspeccnn,qi2017plus,le2018pointgrid}, which present deep learning architectures for tasks such as object classification, object-part segmentation and scene semantic segmentation. These architectures seem to cope well with pointcloud irregular and unorganized nature, achieving impressive results in benchmarking datasets. 
Most of the above methods however impose specific parameterizations such as detecting surface shapes that e.g. afford to walk-on or sit on them, have small categories for affordances and require many examples for scene and objects per category, or work for specific shapes e.g. human actions. Our strategy aims to address all these limitations.

\section{Scalable affordance detection}
\label{sec:affordancedetection}

We here describe our steps and motivation for our proposed approach.

\subsection{Deep learning and affordance data}

Our first attempt to perform affordance predictions consisted in training two deep networks suitable for spatial perception \cite{qi2017plus,qi2016pointnet}, trained for the standard shape classification framework. That is, multiple similar affordance interactions with a single label. We attempted to train the network by presenting it with a pointcloud example and the label of its corresponding affordance. In this sense, our first experiments consisted in training the deep network with 2K examples per class with a total of 85 possible classes (84 affordances and background). However, during extensive experiments we observed that the learning was unable to converge, even with extended data augmentation and parameter tuning. We hypothesize that presenting similar pointclouds with different labels caused the network to \textit{get confused}. For instance, a pointcloud from the edge of a dinning table affords \textit{Placing} objects on top of it but also affords \textit{Sitting} and even \textit{Hanging} objects closer to the edge (e.g a coat-hanger); such pointcloud examples appear to cause the network to get confused in a multi-class single-label scenario. Yet the affordances we wanted to be able to detect are precisely falling in such extended and natural conditions.
The failure of the state-of-the-art, 3D-capable deep network aiming to achieve this task on its own motivated us to consider a hybrid deep-geometric learning approach and leverage the advantages of both approaches.

\subsection{Deep-Geometric scalable affordance detection}
Our work is inspired by both, recent work on geometric affordance detection introduced in \cite{Ruiz2018} which predicted individual affordances from a single example with what the authors call the interaction tensor (iT), and by recent deep learning architectures able to operate in the 3D domain \cite{qi2017plus}. Our work addresses key limitations of the above approaches and contributes in the following significant ways:
\begin{itemize}
\item Introduce a scalable gemetric representation to allow simultaneous multiple-affordance capability in contrast to most affordance methods that learn and work one instance at a time.
\vspace{-3mm}
\item Incorporate a deep 3D saliency mechanism to reduce spatial search and computation time on the original iT. This adopts a data-driven approach in order to mitigate the hand-crafted nature of the iT.
\vspace{-3mm}
\item Significantly outperform baselines in terms of performance and speed, and validate our affordance predictions with human judgment.
\end{itemize}

\subsection{The Interaction Tensor}
\label{subs: iT}
\begin{figure*}[t]
        \centering
        \IfFileExists{./2d_explanation.pdf}{\includegraphics[width=0.9\textwidth]{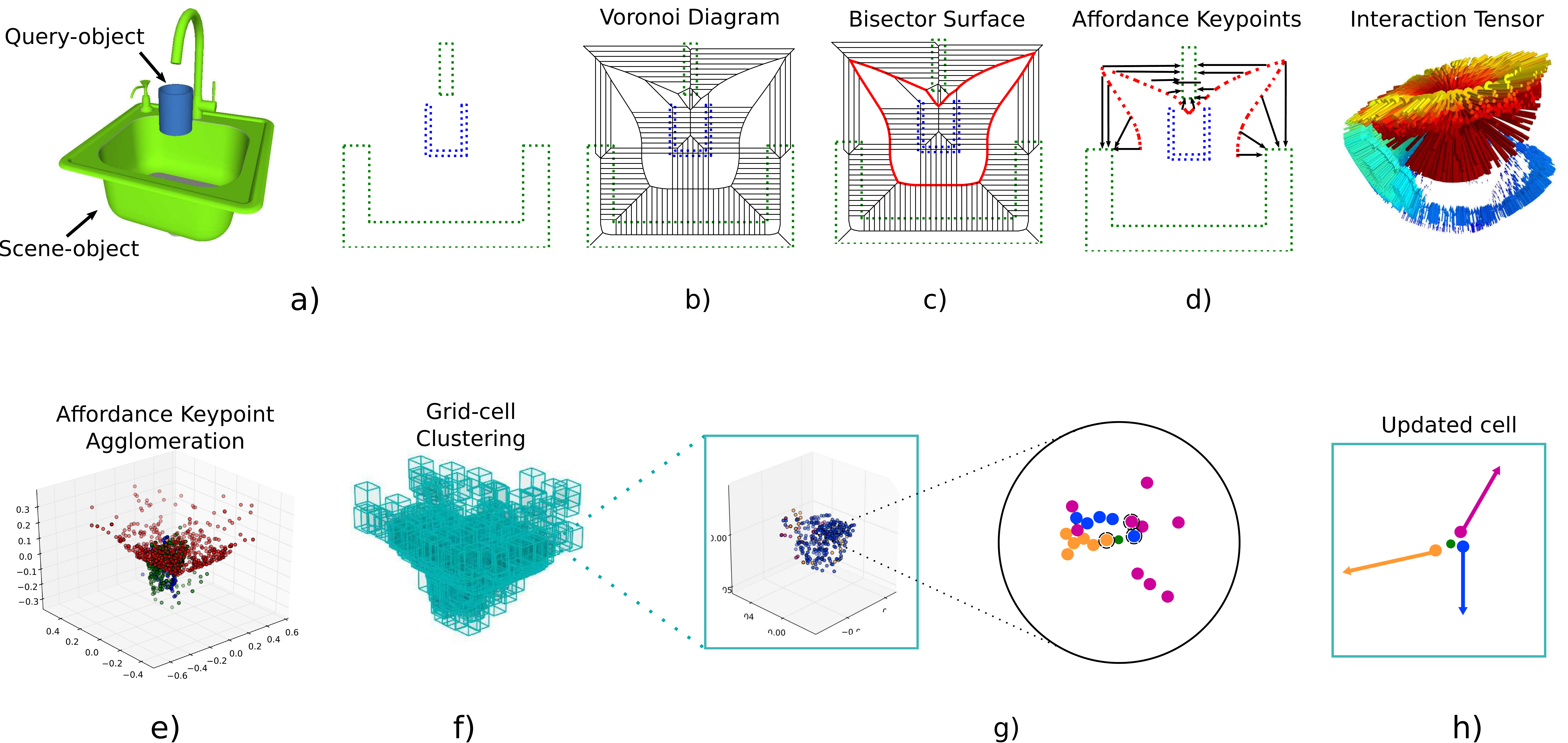}}{\dummyfig{toyInteracionExample}}
      \caption{Top row illustrates how to compute a single interaction tensor for filling a mug a) simplified 2D illustration of interaction, b) voronoi diagram over all points, c) bisector surface shared between interacting objects, d) affordance keypoints and their associated provenance vectors.
      Bottom row shows how to create a scalable and multiple-affordance descriptor from single-example interaction tensors. e) single-affordance keypoints are agglomerated, f) cell grid fitted to the agglomeration, g) one cell can potentially contain keypoints from multiple affordances, h) only closest keypoint (per-affordance) to the cell centroid (green) is taken into account during the update process.}
        \label{fig: single_it}
\vspace{-2mm}
\end{figure*}

The Interaction Tensor (iT) \cite{Ruiz2018} is a vector field representation that characterizes affordances between two arbitrary objects. Using direct, sparse sampling over the iT allows for the determination of geometrically similar interactions from a single \textit{training} example; this sampling comprises what is called {\it affordance keypoints}, which serve to more quickly judge the likelihood of an affordance at a test point in a scene. The iT is straightforward to compute and tolerates well changes in geometry, which provides good generalization to unseen scenes from a single example. The key steps include an example affordance from e.g. a simulated interaction, the computation of a bisector surface between object (query-object) and scene (or scene-object), and estimating provenance vectors, which are the vectors used in the computation of points on the bisector surface. Top row in Fig. \ref{fig: single_it} shows the elements and the process involved in computing an affordance iT for any given objects.

A descriptor for any given affordance is obtained by sampling $N$ \textit{affordance keypoints} (3D point and provenance vector) from the iT example, each one of these keypoints has a weight that encodes the relevance of that particular location for the interaction. The representation is fast to compute and compact since only the affordance keypoints and provenance vectors are used to describe the interaction and neither query, scene or bisector surface are needed afterwards.

\subsubsection{Geometric data augmentation}
As a starting point and similarly to work of Ruiz and Mayol in \cite{Ruiz2018}, we set $N=512$ {\it affordance keypoints}. However, instead of rotating them at test-time to detect affordances at different orientations, we generate a new descriptor by spinning the affordance keypoints (8 orientations in $[0,2\pi)$ around the gravity $\vec{z}$ from the beginning. Then, we zero-mean every new descriptor in order to have all of them sharing the same origin relative to world-frame coordinates. This process leads to 4096 \textit{keypoints} per descriptor.

This is a cheap operation to perform and as we shall see there is no processing impact given our scalable approach.

We carry out this process for multiple affordances using CAD model objects\footnote{https://3dwarehouse.sketchup.com/} of multiple household items from a wide range of geometries and dimensions, we also include a CAD model of a human in order to test \textit{sitting} affordances. With this object collection we investigate scalability and performance. Note that each sample interaction is an affordance on its own right, as it has been established before, this is an intimate relation between object and scene. In our case we have a total of 84 object-affordance pairs of the form \textit{Place-book}, \textit{Hang-umbrella}, \textit{Sit-human}, etc for which an iT descriptor is computed (84 descriptors in total). Some objects afford more than one thing e.g. \textit{Fill-Pitcher} and \textit{Hang-Pitcher}. Notice that it is also possible to consider some top level clustering with conventional generic labels for affordances such as \textit{Placing, Hanging, Filling}. The detailed object-affordance pairs are shown in Fig. \ref{fig: plot_prec}. 

\subsection{iT agglomeration}

Our first step towards scalability is performed in one-shot too. Once all the descriptors have been computed, we agglomerate them in a single pointcloud over which we perform clustering as shown in Algorithm \ref{alg: clustering}. First, we fit a grid of uniform-size cells covering every single \textit{affordance keypoint}. Then, we use as seed-points only the centroid of non-empty cells. For every one of these cells, we only keep the keypoints that are closest to the centroid in a per-affordance basis. For instance, one cell could contain 100 keypoints, all coming from the descriptor of \textit{Placing-bowl}; after the iT clustering process is carried out this cell will only contain the keypoint that is closest to the cell's centroid. Finally, we update the centroid location using the keypoints within each cell, keeping track of the \textit{provenance vectors} associated with them as well as the number of keypoints from each affordance in each cell. 
We attempted more sophisticated ways to learn the agglomeration but found the above straight forward method to be better and faster.

The bottom row in Fig. \ref{fig: single_it} depicts the cell-updating process for iT clustering algorithm (steps 6-9 of Algorithm \ref{alg: clustering})

\begin{algorithm}
\caption{iT clustering}\label{alg: clustering}
\begin{algorithmic}[1]
\Require Affordance keypoints $X=\{x_1,...,x_i\}$, cell size $e$
\Ensure Cluster centroids $C=\{c_1,...,c_j\}$
\State Initialize C with centroids evenly distributed in $[x_{\text{min}},x_{\text{max}}]$ according to $e$.
\State Assign $x_i$ to cluster $argmin_j \lVert x_i- C_j\rVert_2$ $\forall x_i \in X$
\State Remove empty clusters
\State Initialize update sets ${Y_1,...,Y_j}$ to empty
\ForAll{Clusters $C$}
  \ForAll{Affordances $A=\{a_1,...,a_k\}_{\neq}$ in $C_j$}
  \State Recover all $x$ from affordance $a_k$ 
  \State Assign $argmin_i \lVert x^k_i- C_j\rVert_2$ to $Y_j$ 
  \EndFor
\EndFor
\State Update centroids: $c_j \gets \frac{1}{\vert Y_j \vert}\sum_{y\in Y_j}y$
\end{algorithmic}
\end{algorithm}

The clustering process leads to a reduced number of 3D points (cell centroids) that represent a large number of affordance keypoints. This reduced number of new keypoints and their associated \textit{provenance vectors} are used to compute and predict affordance candidate locations at test time.

\section{Learning to sample from scene saliency}
\label{section:saliency}
The iT approach relies on affordance keypoints that are sampled from a tensor in order to form a descriptor. Although a sparse, empirically-found sample size has been proposed in \cite{Ruiz2018}, we investigate and provide insight into the optimal affordance keypoint sampling by leveraging state-of-the-art deep learning methods. For this, we capitalize on the PointNet++ architecture \cite{qi2017plus}, which has shown noticeable results for shape classification on 3D data. This architecture is advantageous in our problem mainly for two reasons: 1) its ability to capture local structures induced by the metric space points live in, and 2) its robustness against non-uniform sampled pointclouds. We employ the network  to learn the specific locations in the input pointcloud (scene) that are used for prediction.  This is achieved by keeping track of the points that activate neurons the most. These points suggest salient 3D locations in a given pointcloud. This is analogous to the concept of critical pointsets presented in \cite{qi2016pointnet}, which are the 3D points that contribute to the max-pooled features in the network's first level filters. 

\subsection{Saliency from affordance predictions}

Ideally we want to learn an affordance representation that efficiently accounts for the presence of multiple interactions in a given pointcloud. With this goal in mind
we frame the learning task as multi-label, multi-class classification. In this respect, the training label of a given pointcloud should include all the possible interactions that the current example affords. In order to obtain such training data, we post process the predictions made by the iT algorithm (single-label); since now we need to present the network with prediction examples (e.g. pointclouds) shared by multiple affordances. Whereas this process could be regarded as trivial, it should be noted that the iT agglomeration approach does not need it. Clustering affordance keypoints from individual interactions (from a single \textit{training} example) allows us to predict multiple affordances in any given pointcloud, as shall be shown later.

For our setup, we replace the softmax layer output in PointNet++ with a sigmoid layer and train with cross-entropy $L=-\sum_{i=1}^k y_ilog(\hat{p_i})+(1-y_i)log(1-\hat{p_i})$ with $k=85$ (84 affordances and background). Additionally, we perform $L_2$-norm regularization since over-fitting was observed during preliminary experiments. In order to train the network we zero-mean the pointclouds, which allows us to track scene saliency for different pointclouds relative to the same reference frame. Most shape classification approaches normalize the training data to a unit-sphere or unit-box; however, this is not feasible in our learning approach given that we work with real-world scales. In other words, having a pointcloud of a chair of 1 meter-height is substantially different from a \textit{toy chair} with a height of 10 cm; the latter would not afford \textit{Sitting} for a human. For this reason, and in contrast with \cite{qi2017plus}, we allow the sampling regions to change proportionally to the current training pointcloud. These regions are used by the network to learn features at different hierarchies and pointcloud densities.

Due to the fact that the iT relates points in the two interacting objects (i.e. scene and query-object), we can \textit{easily} project salient points (from the scene) learned by the network back into their associated iT location.
Briefly speaking, for all scene-salient points we compute the nearest-neighbour in the iTs of all the interactions afforded by the current pointcloud. This is the inverse process to the one shown in Fig. \ref{fig: single_it}.d. Given that the iTs are very dense entities we use a grid representation (i.e. cell grid) to alleviate the back-projection process (blue cells in Fig. \ref{fig: method}: Multiple-affordance representation). Once all salient locations have been projected into their associated iT, we create a new multiple-affordance descriptor by considering the locations in the iT agglomeration (i.e. cells) that \textit{received} projections the most.

\section{Experiments and Evaluation}
\begin{figure*}[ht]
         \centering
         \IfFileExists{./test_time.pdf}{\includegraphics[width=0.9\textwidth]{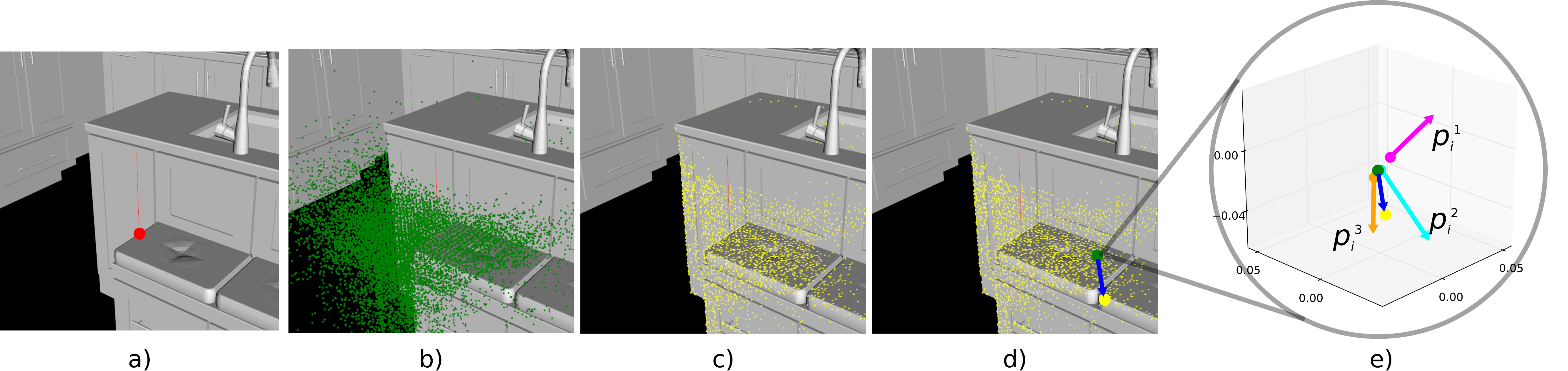}}{\largedummyfig{Figure needs to show test time vectors and with keypoints inside cell}}
       \caption{Illustration of affordance prediction at test-time. a) a test-point is sampled from the input scene (red), b) the agglomerative representation (green) is aligned relative to this test-point, c) The closest scene point (yellow) for every centroid in the agglomeration, d) an example test-vector (blue) from a cell centroid to its closest scene point, e) test-vector is compared against the stored provenance vectors $p_i^k$ associated with affordance keypoints in that cell. In this particular cell, 3 scores are obtained.}
         \label{fig: testtime}
\vspace{-2mm}
\end{figure*}
\label{section:evaluation}
We are interested in giving answer to the perceptual question of "What can I do here?". We here predict up to 84 affordances or interaction possibilities in any given location of an input scene. This is achieved by testing multiple locations all over the scene. Whereas this could be seen as an exhaustive process, it should be reminded that our approach does not make assumptions about appearance or complex surface features on the elements or parts in the scene that afford the interactions of interest. 
In order to make predictions we: i) randomly sample a test-point in the input scene, ii) perform 1-nearest-neighbor search for every keypoint in the agglomerative descriptor using the voxel surrounding the test-point , iii) estimate \textit{test vectors} and compare against \textit{provenance vectors} to produce a score using Eq (\ref{eq:score}).
\begin{equation}\label{eq:score}
s^k= \frac{1}{N^k}\sum_{i=1}^{N^k}\frac{1}{\sqrt[]{2\pi (w^k_i)^2}}e^{-\frac{(\Delta^k_i)^2}{2 (w^k_i)^2}},
\end{equation}
where
\[
\Delta^k_i= \frac{\lVert \vec{v_t}_j - \vec{p}^k_i \rVert}{\lVert \vec{p}^k_i \rVert},
 \]
and where $w^k_i$ is the keypoint's weight obtained from the magnitude of its corresponding \textit{provenance vector} $p^k_i$ ($i-$th keypoint of affordance $k$). $\Delta_i^k$ is the difference between test vector $\vec{v_t}_j$ (estimated using the $j$-th cell-centroid) and provenance vector $p^k_i$. Equation (\ref{eq:score}) fits a Gaussian distribution to the difference between vectors (magnitude and orientation), where the acceptable variance is inversely proportional to the keypoint’s weight $w^k_i$.
Fig. \ref{fig: testtime} depicts the steps followed at test-time in order to predict affordance candidate locations. 

For training the saliency network we use the scene dataset presented in \cite{Ruiz2018} comprised by pointclouds of indoor environments from synthetic data and real RGBD scans. After processing the scens, the affordance dataset is comprised by 918K pointclouds (10K per-affordance on average) with a 80/20 split for training and validation. Data augmentation is performed on-line by rotating the pointclouds around the vertical axis, adding jitter and randomizing the points sampled at the input. Performance of the adapted PointNet++ architecture, when tested on its own, are shown as "m-PointNet++" throughout the following subsections. We test and show examples of our predictions on 150 ScanNet \cite{scannet2017} scenes (randomly selected) comprising living rooms, kitchens, dinning rooms and offices. Qualitative results of our predictions are shown in Figs. \ref{fig: first_image_examples}, \ref{fig: method} and \ref{fig: rgbd_examples} throughout the paper. 
Our code and data is available at the authors' public repository.

\subsection{Optimal detections}
\label{aturk}
\begin{figure}[ht]
        \centering
        \IfFileExists{./roc_amazon2.pdf}{\includegraphics[width=0.45\textwidth]{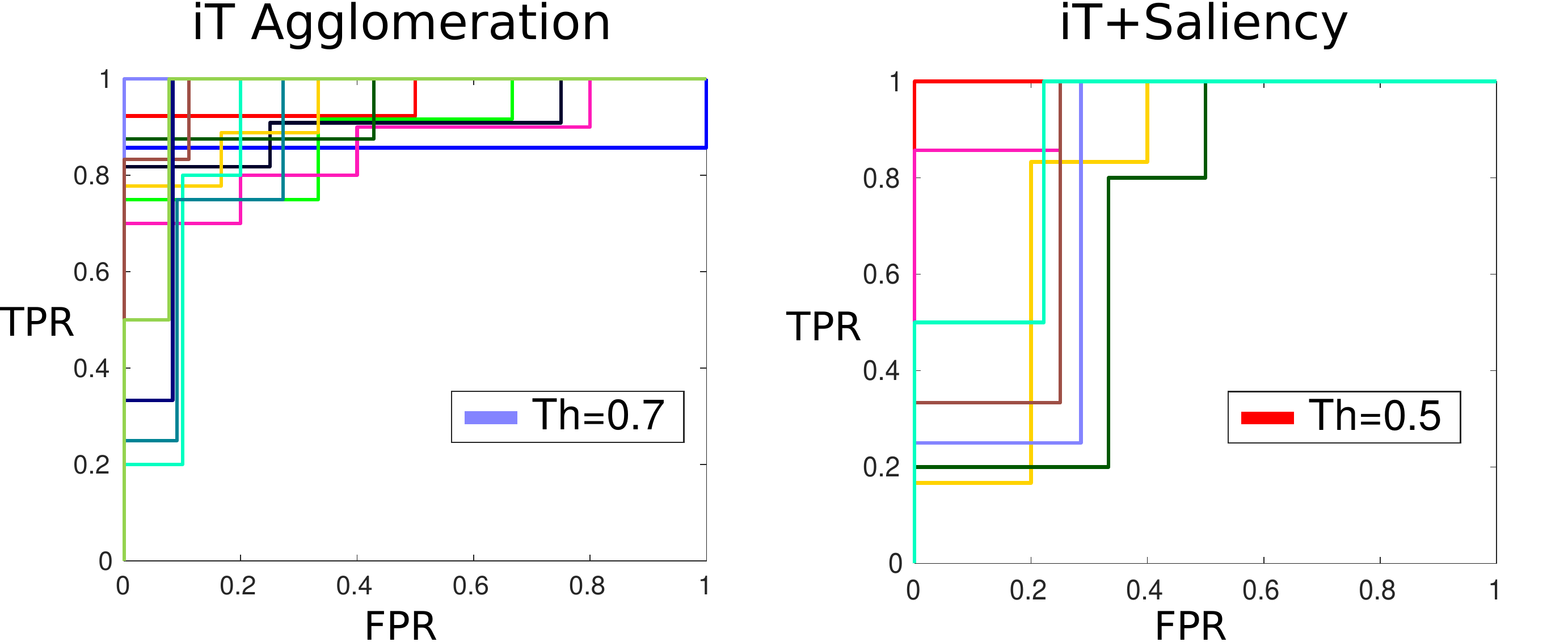}}{\dummyfig{ROCs from amazon turk}}
      \caption{Mechanical turk evaluation. ROC plots show the family of classifiers generated by setting different thresholds bands to the prediction score. Both approaches show a similar performance: the affordance predictions with a score above their respective threshold are deemed as good candidates according to humans every time.}
        \label{fig: ROCs}
\vspace{-2mm}
\end{figure}

Affordances by nature are elusive to groundtruth without subjective judgment of the likelihood of an interaction. In our work affordance location predictions are made by setting a threshold to the output (score) of the algorithm. In order to determine the threshold that produces the best results we use Amazon Mechanical Turk; where we ask people to evaluate the predictions made with our algorithms based on the smallest cell sizes. A total of 4.8K example predictions representing different scores were shown to 69 humans evaluators (turkers). These subjects had to select a "winner" from two possible options showing the same affordance-object pair resulting from different scores. Using these pairwise human judgments, we fit a Bradley-Terry model \cite{Bradley1952} to compute the "true" ranking of human evaluations; with this ranking we asses the performance of our algorithm. Fig. \ref{fig: ROCs} shows the family of classifiers induced by setting different threshold values at the score of the iT agglomeration and saliency-based iT algorithms. 
In this figure can be seen that both methods achieved good performance according to human criteria yet the optimal thresholds are different. The method based on agglomeration and clustering of iT descriptors needs a threshold of 0.7 in order to produce prediction that agree with human criteria; on the other hand, the saliency-based method performs similarly with a threshold of 0.5. This is related to the fact that, as seen in the following subsection, our saliency-based method achieves higher precision rates; meaning that we can relax the threshold without compromising the quality of the predictions.

\subsection{Individual vs Multiple predictions}
\label{sec:metrics}
\begin{figure*}[ht]
        \centering
        \IfFileExists{./precisions_all.pdf}{\includegraphics[width=0.95\textwidth]{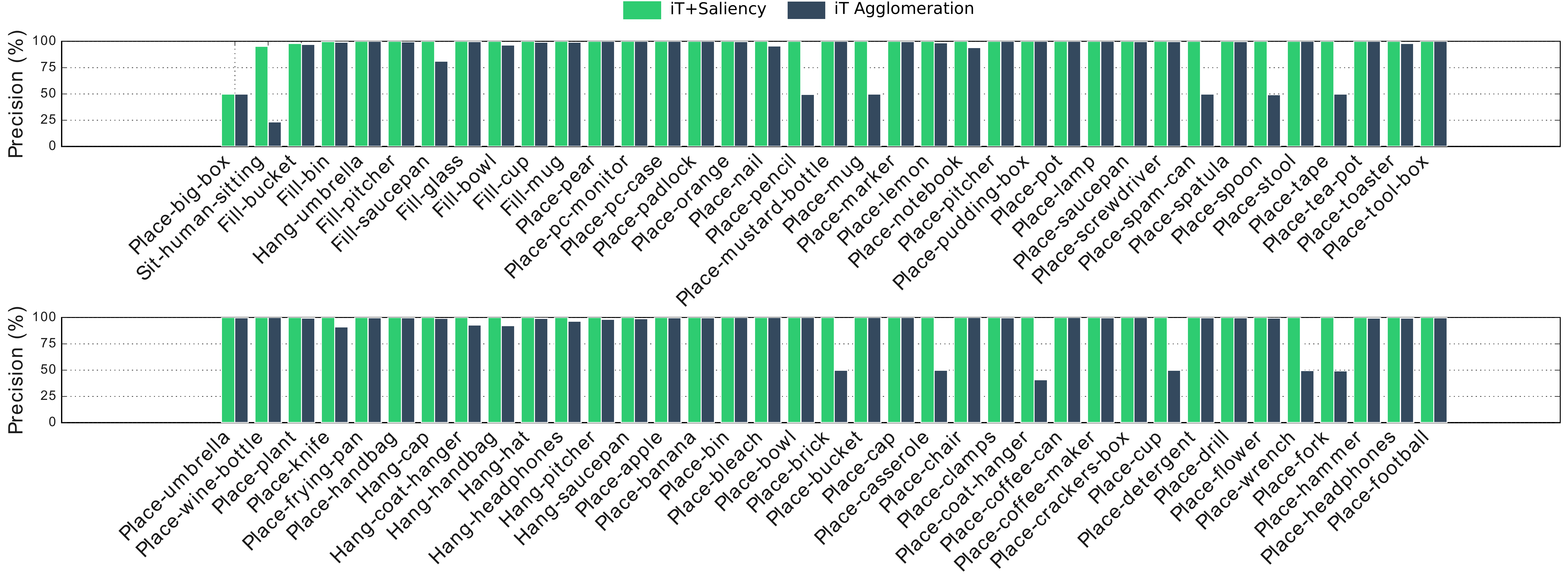}}{\largedummyfig{Precision plots for individual affordances}}
      \caption{Precision achieved with multiple-affordance representations of cell size 0.5cm. Important differences are noted in \textit{Placing} affordances which were largely regarded as the \textit{easiest} interactions. Interestingly both approaches struggle the most with \textit{Placing big-box}, which has several short-length vertical provenance vectors (under the box) difficult to match during testing.}
        \label{fig: plot_prec}
\vspace{-2mm}        
\end{figure*}

By now we have demonstrated that the predictions made with our algorithm align well with what humans expect to afford in a given scene/location. In an effort to further asses the performance of our algorithms we compare our multiple affordance predictions against those produced in a single-affordance scenario.

These ''baseline`` predictions are obtained by individually testing all affordances using the single-affordance prediction algorithm, as in \cite{Ruiz2018}, for every affordance-object pair in our study. We treat these as "ground-truth" in order to compute performance metrics for our predictions.

\begin{table}[ht]
\centering
\resizebox{0.3\textwidth}{!}{%
\begin{tabular}{@{}ccc@{}}
Method & \begin{tabular}[c]{@{}c@{}}Cell-size\\ {[}cm{]}\end{tabular} & AUC \\ \midrule
\multirow{3}{*}{iT Agglomeration + Saliency} & 0.5 & \textbf{0.6816} \\
 & 1 & 0.4588 \\
 & Single & 0.2722 \\ \midrule
\multirow{3}{*}{iT Agglomeration} & 0.5 & 0.5467 \\
 & 1 & 0.3043 \\
 & iT-All & 0.3102 \\ \midrule
m-PointNet++ &  & 0.1879
\end{tabular}%
}
\caption{Average performance of the methods for multiple affordance detection in terms of Area Under the PR Curve (AUC). 
}
\label{tab: auc}
\vspace{-2mm}
\end{table}
For the comparisons we include two additional versions of our algorithms. In the case of iT agglomerations we tested keeping all the keypoints (and their associated provenance vectors) inside the cell during clustering. This is shown as ``iT-All" and the intention of this was to investigate the possible loss of information caused by only considering the closest per-affordance keypoint inside each cell. For the case of our saliency-based method, we consider an alternative approach where salient locations are learned individually per affordance and then their corresponding keypoints are agglomerated to produce a mulitple-affordance representation. This saliency-based alternative is shown as ''Single``. We also show the performance achieved by the modified PointNet++ architecture used to learn saliency (shown as  m-PointNet++) when tested on its own. Table \ref{tab: auc} shows the average performance of the methods being investigated, where it can be observed that saliency-based method showed overall better performance.

\begin{figure}[ht]
        \centering
        \IfFileExists{./prec_recall3.pdf}{\includegraphics[width=0.48\textwidth]{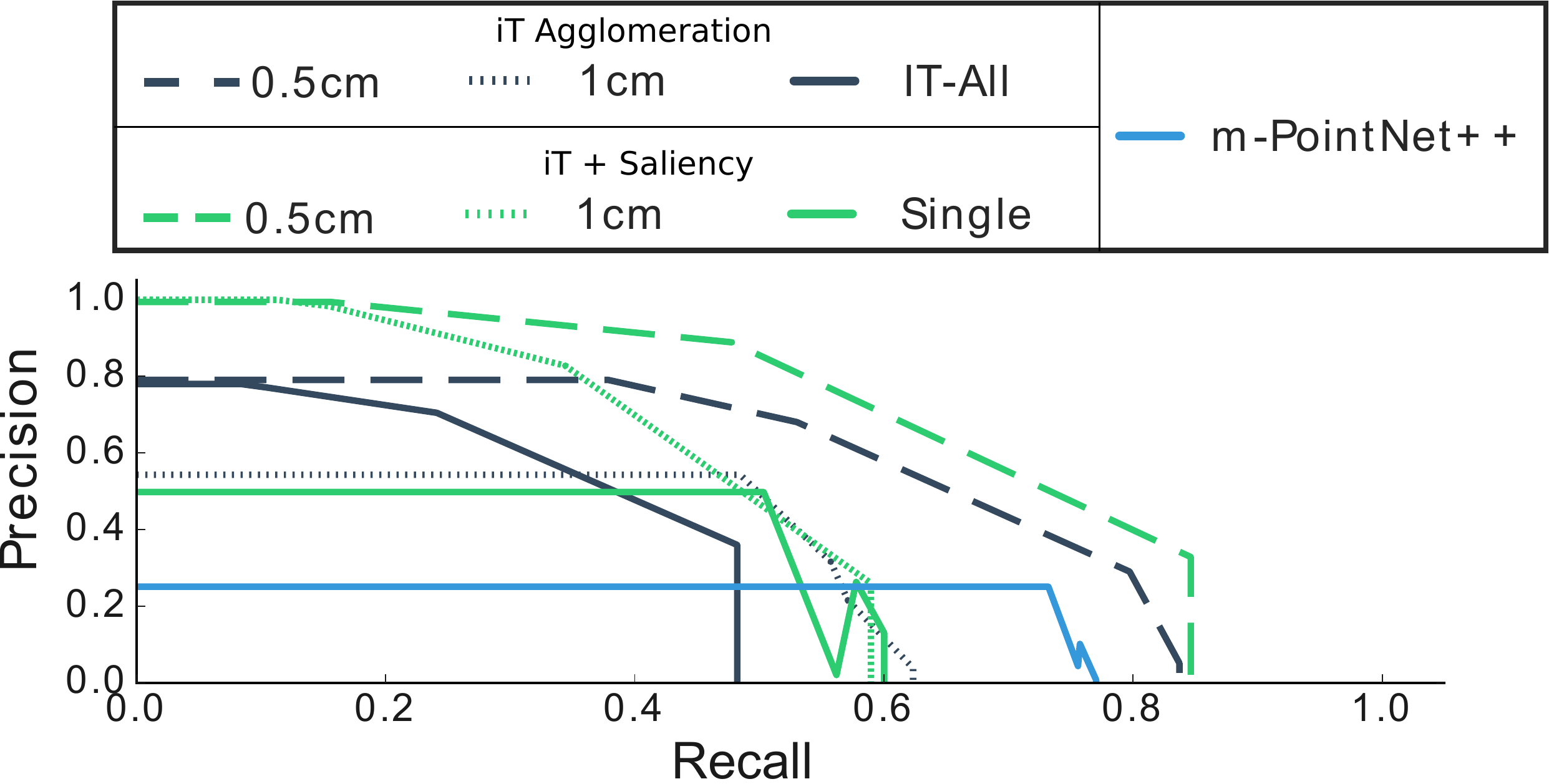}}{\dummyfig{Pop cells plot, could be a table as in PointGrid}}
      \caption{Precision-Recall curves of our methods when compared with predictions made by single-affordance algorithm. Observe that precision drops to zero at specific recall values, this is associated to quantization error introduced by cell clustering. }
        \label{fig: prec-rec}
\vspace{-3mm}
\end{figure}

It should be noted that there is a important imbalance in affordance data. For instance, consider a kitchen environment for \textit{Filling} affordances; there is usually one location that truly affords these interactions: the faucet/tap. In this scenario almost every single location is a negative example for this affordance, by always predicting "background" we could achieve very high accuracy. For this reason we evaluate with precision-recall metrics, Fig.\ref{fig: plot_prec} and \ref{fig: prec-rec} show the precision and recall values achieved with our algorithms.

\begin{figure}[ht]
        \centering
        \IfFileExists{./cell_sizes3.pdf}{\includegraphics[width=0.48\textwidth]{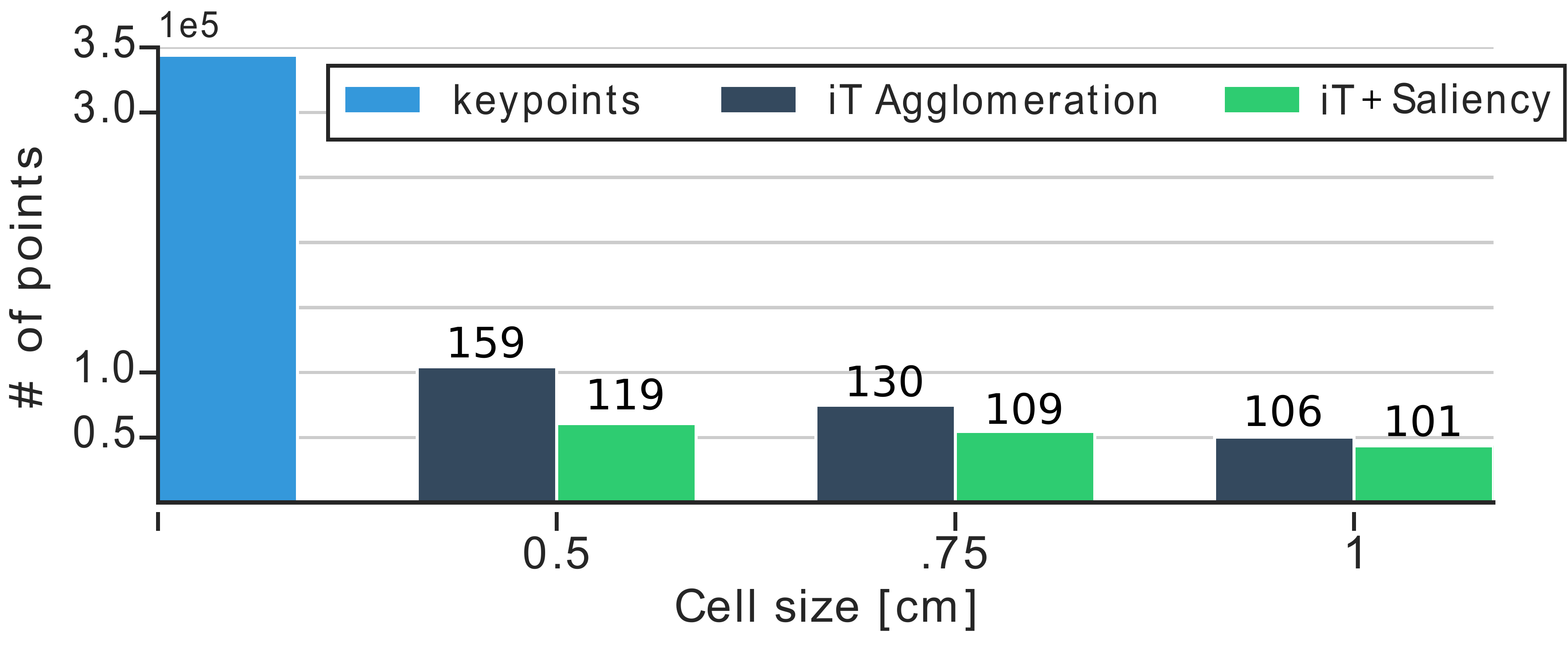}}{\dummyfig{Pop cells plot, could be a table as in PointGrid}}
      \caption{Bar plot shows the dimentionality reduction achieved with our methods for different cell sizes. We reduce up to 6 times the number of keypoints required to make predictions. Numbers above each bar show prediction time (milliseconds) per test-point of the input scene.}
        \label{fig: sizes}
\vspace{-3mm}
\end{figure}

\begin{figure*}[ht]
        \centering
        \IfFileExists{./rgbd_examples_real_kitchen1.pdf}{\includegraphics[width=0.95\textwidth]{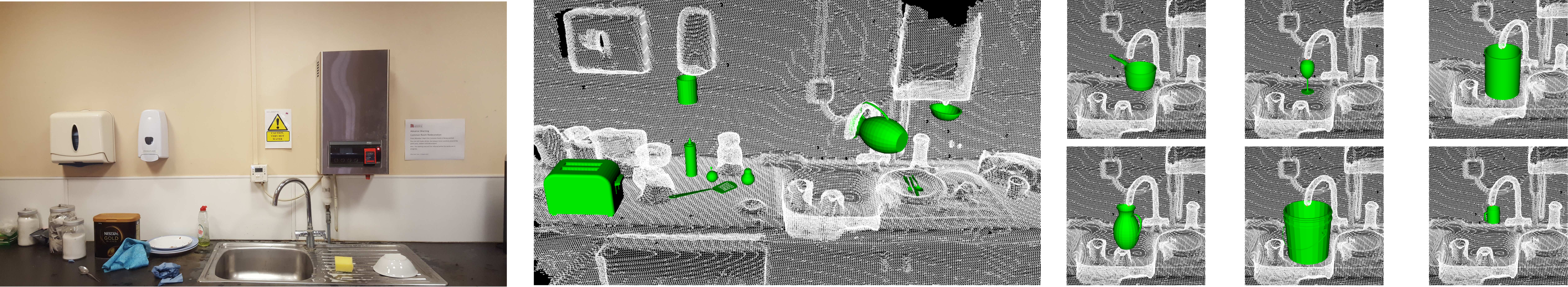}}{\largedummyfig{RGBD examples}}
      \caption{Multiple interactions afforded by the same pointcloud (tap/faucet) predicted in a never seen before scene.}
        \label{fig: rgbd_examples}
\vspace{-3mm}
\end{figure*}

It is worth mentioning that the iT agglomeration method performed favourably even when a single \textit{training} example is considered. Then, by combining the interaction tensors with the salient locations learn by the network we are able to improve the performance and, as shall be shown later, more quickly asses the interaction opportunities of a given location in the scene. The deep network on its own denoted a rather poor performance; when compared with single affordance predictions the network was outperformed by all other methods. We think this highlights the importance of the geometric features that the iT is able to describe. Fig. \ref{fig: plot_prec} shows per-affordance precision achieved by our top performing methods.

From Fig. \ref{fig: prec-rec} is worth noting that none of our methods achieves 100\% recall, this means that we are not able to recover or predict every possible affordance in every location. This is the compromise we make in order to perform fast and multiple-affordance predictions. In other words, the methods here presented are optimal if the task is to quickly evaluate the affordance possibilities at any given location with a high precision; but, if the task is to retrieve all possible "combinations" or every affordance that exists across all the scene while speed is not crucial, performing single-affordance predictions is perhaps a better approach. In spite of this, we show next that the our predictions are equivalent to those produced by the single affordance predictions.

\textbf{Human evaluation} \enskip We asses the predictions made with our algorithms by asking human evaluators to select from two options the one that best depicted the intended interactions. These options consisted on: the top predictions made by the single affordance algorithm and the top predictions of our multiple-affordance method, shown in a per-affordance basis. Additionally, among the options shown to people we included top prediction made with a \textit{naive} baseline method that uses Iterative Closest Point (ICP) \cite{Rusu2011}. This baseline computes a score from the best alignment (i.e. rigid transformation) between a target pointcloud (interaction training example) and the pointcloud being tested. A total of 1200 pair-wise comparisons were shown to 48 turkers. We found that 48\% of the time people chose the multiple-affordance predictions as the best option when compared against single-affordance predictions. Note that a random guess is 50\%, which means that our predictions are regarded as good as the "ground truth" (i.e. single affordance predictions). Most of the time people were not able to distinguish the interactions generated by our method from those produced by single predictions. On the other hand, when compared with the ICP baseline, our predictions were chosen 87\% of the time. 

\subsection{Frame rates and quantization}
We evaluate the effect that the size of cells in the grid has in terms of speed or prediction time for the methods that we propose. Our first consideration for clustering explored non-uniform spatial representations such as those in e.g. OcTrees; however, due to the variance in sparsity and size of the affordances we study, that representation did not perform as well as sparse but uniform-sized cells.  All the tests are carried out on a desktop PC with a NVIDIA Titan X GPU. Fig. \ref{fig: sizes} shows the dimensionality of our multiple-affordance representation and the prediction rates according to the cell size. Looking at this figure it stands out the large reduction that we are able to achieve with our proposed approaches: both of them reduce the number of points in the representation by nearly 6 times (344K vs 60K keypoints). The prediction rates on the same figure show that using grids with a cell size of 1 $\text{cm}^3$ allows us to detect up to 84 affordances at 10 different locations per iteration on the input scene. This is significantly faster (7x improvement) than predicting affordances by trying 84 descriptors at test time, which would require 840 ms per test-point (average of 10 ms per affordance as reported in \cite{Ruiz2018}). Due to the fact that our prediction algorithm performs a NN-search in order to estimate test-vectors and compare them against provenance vectors, the complexity of such operation depends heavily on the dimension of the multiple-affordance representation (i.e. number of centroids/keypoints). More points in the representation require more computations; therefore, reducing the representation allows us to perform faster evaluations at test-time. Even with such reduction in dimensionality our method, as shown earlier, is able to produce top quality affordance predictions. Fig \ref{fig: rgbd_examples} and the suplementary material offer further examples of our predictions that show generality and multiple affordance estimation per scene.
\vspace{-3mm}

\section{Conclusions}
\label{conclusions}
We have developed and evaluated a scalable, real-time approach for multiple affordance prediction. Our approach leverages advantages of a compact geometric representation with scene saliency by a deep learning architecture. We predict up to 84 affordances on over 150 real previously unseen scenes and in a way that aligns well within the intrinsically subjective nature of affordances as validated with crowd-sourced human judgment. In such evaluation, our output is preferred 87\% of the time. Furthermore, we show high rates of improvement with almost four times better performance over a deep-learning-only baseline and a 7 times faster operation compared to previous art. As a result of our detection rates, we see many venues for its application such as in semantic scene understanding, robot planning and augmented reality where scenes can be augmented using discovered affordances rather than pre-scripted as is usually the case. We are also further interested in extending this work to incorporate dynamic elements of how affordances develop in a given space. Overall we see affordance determination as a key competence for Vision to gain insight about the world and methods that aim to use few examples and are able to generalize from there, as the best north stars in this space.

{\small
\bibliographystyle{ieee}
\bibliography{biblio}
}
\end{document}